\documentclass[sigconf]{acmart}
\usepackage{url}
\usepackage{tabularx}
\usepackage{booktabs}
\usepackage{ragged2e}
\AtBeginDocument{%
  }

\copyrightyear{2026}
\acmYear{2026}
\setcopyright{cc}
\setcctype{by}
\acmConference[SCA/HPCAsia 2026]{SCA/HPCAsia 2026: Supercomputing Asia and International Conference on High Performance Computing in Asia Pacific Region}{January 26--29, 2026}{Osaka, Japan}
\acmBooktitle{SCA/HPCAsia 2026: Supercomputing Asia and International Conference on High Performance Computing in Asia Pacific Region (SCA/HPCAsia 2026), January 26--29, 2026, Osaka, Japan}
\acmPrice{}
\acmDOI{10.1145/3773656.3773682}
\acmISBN{979-8-4007-2067-3/2026/01}

\begin{document}

\title[AMR-EnsembleNet]{Fusing Sequence Motifs and Pan-Genomic Features: Antimicrobial Resistance Prediction using an Explainable Lightweight 1D CNN - XGBoost Ensemble}

\author{Md. Saiful Bari Siddiqui}
\email{saiful.bari@bracu.ac.bd}
\orcid{0009-0000-7781-0966}
\affiliation{%
  \institution{BRAC University}
  \city{Dhaka}
  \country{Bangladesh}
}

\author{Nowshin Tarannum}
\email{nowshin@iub.edu.bd}
\orcid{0000-0003-3897-4516}
\affiliation{%
  \institution{Independent University Bangladesh (IUB)}
  \city{Dhaka}
  \country{Bangladesh}}

\renewcommand{\shortauthors}{Siddiqui and Tarannum}

\begin{abstract}
Antimicrobial Resistance (AMR) is a rapidly escalating global health crisis. While genomic sequencing enables rapid prediction of resistance phenotypes, current computational methods have limitations. Standard machine learning models treat the genome as an unordered collection of features, ignoring the sequential context of Single Nucleotide Polymorphisms (SNPs). State-of-the-art sequence models like Transformers are often too data-hungry and computationally expensive for the moderately sized datasets that are typical in this domain. To address these challenges, we propose \textbf{AMR-EnsembleNet}, an ensemble framework that synergistically combines sequence-based and feature-based learning. We developed a lightweight, custom 1D Convolutional Neural Network (CNN) to efficiently learn predictive sequence motifs from high-dimensional SNP data. This sequence-aware model was ensembled with an XGBoost model, a powerful gradient boosting system adept at capturing complex, non-local feature interactions. We trained and evaluated our framework on a benchmark dataset of 809 \textit{E. coli} strains, predicting resistance across four antibiotics with varying class imbalance. Our 1D CNN-XGBoost ensemble achieved top-tier performance across all the antibiotics, reaching an accuracy of 0.963 for Ciprofloxacin (CIP) and the highest Macro F1-score of 0.691 for the challenging Gentamicin (GEN) AMR prediction. We also show that our model consistently focuses on SNPs within well-known AMR genes like \textit{fusA} and \textit{parC}, confirming it learns the correct genetic signals for resistance. Our work demonstrates that fusing a sequence-aware 1D CNN with a feature-based XGBoost model yields a powerful ensemble that overcomes the limitations of either an order-agnostic or a standalone sequence model. Our codes are publicly available on GitHub at: \url{https://github.com/Saiful185/AMR-EnsembleNet}.
\end{abstract}




%\received{19 September 2025}
%\received[accepted]{20 October 2025}

\maketitle

\section{Introduction}

Antimicrobial Resistance (AMR) represents a critical and escalating threat to global health. Often called the "silent pandemic," the rise of drug-resistant pathogens jeopardizes decades of medical progress, threatening the safety of routine procedures from surgery to chemotherapy. The World Health Organization estimates that without significant intervention, AMR could cause 10 million deaths annually by 2050, making it a leading cause of mortality worldwide \cite{WHO_AMR_Report, ONeill_Review}. A major factor fueling this crisis is the slow pace of diagnostics. Traditional culture-based methods take 2-3 days to identify an effective antibiotic, forcing clinicians to rely on broad-spectrum drugs that can inadvertently promote further resistance.

The genetic basis of AMR is the core reason why sequencing is a viable alternative to culture-based methods. Resistance phenotypes are fundamentally driven by changes in a bacterium's DNA. These changes can range from single nucleotide mutations in a drug's target protein, to the acquisition of entire genes that encode drug-inactivating enzymes \cite{Sherry2025Genomics}. Whole Genome Sequencing (WGS) provides a complete readout of a pathogen's genetic makeup, capturing all this potential variation. WGS thus offers a powerful alternative, able to deliver a complete resistance profile from a pathogen's DNA in a matter of hours. However, translating this vast genomic data into clinically actionable predictions remains a significant computational challenge. The most common approach involves searching for a known list of resistance genes \cite{Gygli2019WGS}. While useful, this "gene hunting" method is fundamentally limited by existing knowledge and cannot identify novel or emerging resistance mechanisms. Another alternative is using Single Nucleotide Polymorphisms (SNPs). They can be used to create a high-resolution map of the genetic differences that distinguish resistant strains from susceptible ones.

To move beyond this limitation, researchers have applied standard machine learning models like Random Forest and XGBoost to pan-genomic data, such as Single Nucleotide Polymorphism (SNP) matrices~\cite{Drouin2016Predictive},~\cite{Sakagianni2023MLAMR},~\cite{Li2018Genomic}. These methods have shown great promise but have a critical drawback: they treat the genome as an unordered collection of features, ignoring the valuable information contained in the sequence and local context of the SNPs. This is a significant omission, as mutations within a single gene or on a mobile genetic element often work together to confer resistance~\cite{Shi2019DeepFeature}.

While state-of-the-art architectures like Transformers are designed to understand sequential data, their success in other fields often relies on pre-training with massive datasets \cite{Devlin2019BERT},~\cite{Kaplan2020Scaling}. For tasks like AMR prediction, however, such large-scale, domain-specific pre-trained models are not readily available. Training a Transformer from scratch is notoriously data-hungry. Its depth and large number of parameters make it difficult to train on the typically moderately-sized genomic datasets without a significant risk of overfitting~\cite{Zhang2023TransformersBioinfo},~\cite{siddiqui2024d2c}. This leaves a crucial gap in the field: a need for a model that can effectively learn the dynamics of the genome from a limited dataset, without the prohibitive data and computational costs of a full Transformer.

In this work, we bridge this gap by developing a synergistic ensemble framework. We pair a lightweight 1D Convolutional Neural Network (CNN), designed to efficiently learn local SNP motifs, with an XGBoost model that captures complex, non-local feature interactions. We show that this hybrid approach leads to a more robust and accurate predictor. The main contributions of this paper are:
\begin{itemize}
    \item The development of a lightweight 1D CNN architecture that effectively models local genomic context from high-dimensional SNP data.
    \item The demonstration that a synergistic ensemble of that 1D CNN architecture and XGBoost, \textbf{AMR-EnsembleNet}, outperforms individual baseline models, especially on imbalanced datasets.
    \item Attaining high predictive accuracy on a benchmark \textit{E. coli} dataset, while demonstrating a notable improvement compared to strong machine learning baselines in recall for the most clinically relevant, imbalanced antibiotic cases.
    \item Providing an interpretable framework that highlights the most predictive SNP features, which were subsequently mapped to known AMR-associated genes, demonstrating the biological relevance of the model.
\end{itemize}

\section{Theoretical Background}

\subsection{The Genetic Basis of Antimicrobial Resistance}

Antimicrobial resistance (AMR) is the ability of microorganisms to resist the effects of drugs that are designed to kill them or inhibit their growth. In bacteria, resistance to antibiotics can be intrinsic or acquired. Intrinsic resistance results from their inherent physiological or structural characteristics, such as the presence of an impermeable outer membrane in Gram-negative bacteria or multi-drug efflux pumps that expel antibiotics from the cell \cite{Tunstall2020}. On the other hand, acquired resistance requires a genetic mutation or the acquisition of new genetic material. Bacteria can acquire resistance genes from other bacteria via mobile genetic elements (MGEs) like plasmids, a process known as horizontal gene transfer (HGT), or they can undergo spontaneous mutations \cite{Coculescu2009}. While HGT is a major driver of resistance spread, mutational changes, particularly single nucleotide polymorphisms (SNPs), are fundamental to the evolution of new resistance mechanisms.

Single nucleotide polymorphisms, or SNPs, are the most common type of genetic variation, involving a change to a single nucleotide at a specific position in the genome. While many SNPs are neutral, those occurring in critical locations can alter the function of genes, leading to phenotypic changes like antibiotic resistance \cite{Shi2019DeepFeature}. These point mutations can introduce resistance by modifying the antibiotic’s molecular target, altering the expression of genes involved in drug transport, or activating alternative metabolic pathways that bypass the drug’s action \cite{Coculescu2009},~\cite{Tunstall2020}.

An SNP can directly impact a gene by changing the amino acid sequence of the protein it codes for, which is known as a non-synonymous mutation. This can alter the protein’s shape or function. For example, a key SNP in a gene encoding a ribosomal protein could prevent an antibiotic like streptomycin from binding to the ribosome, thereby rendering the drug ineffective \cite{Tunstall2020}. Other SNPs can occur in regulatory regions, such as promoters, which can increase or decrease gene expression. A mutation in a promoter region could, for instance, lead to the overexpression of a drug efflux pump, allowing the bacterium to expel an antibiotic before it can reach a toxic concentration \cite{Coculescu2009}.

The link between specific SNPs and antibiotic resistance in \textit{E. coli} is well-established. A classic example is resistance to fluoroquinolones, a major class of antibiotics. These drugs work by disabling two essential enzymes, DNA gyrase and topoisomerase IV, which are critical for the bacterium's DNA replication. Resistance commonly arises from specific SNPs in the \textit{gyrA} and \textit{parC} genes, the very genes that provide the instructions for these enzymes. For instance, a single nucleotide change that causes the amino acid serine to be replaced by leucine at position 83 of the GyrA protein is known to physically block the antibiotic from docking with the enzyme, rendering the drug ineffective \cite{Tunstall2020}. Similarly, studies have identified single point mutations in genes like \textit{tetA} that are sufficient to confer resistance to other antibiotics like tetracycline \cite{Boripun2023}. These examples show that even a single-letter change in the genetic code can have profound functional consequences.

Discovering these critical SNPs is a key challenge. Traditional genetic methods, like PCR assays, are analogous to using a "find" function with a predefined search term; they can only detect the specific resistance genes or mutations that we already know to look for. This approach is fundamentally limited and cannot identify new or unexpected causes of resistance. The development of Whole Genome Sequencing (WGS), however, has been a game-changer \cite{Ren2022}. Instead of searching for a few known markers, WGS provides the entire genetic sequence of a bacterium. This allows for an unbiased, comprehensive survey of every genetic variation, including all SNPs, across the whole genome, providing a complete dataset from which to learn the patterns of resistance \cite{Shi2019DeepFeature}.

\subsection{From SNPs to Prediction: Machine Learning Approaches}
Once all relevant SNPs from a bacterial population are identified and aligned to a reference genome, the result is a high-dimensional SNP matrix. In this matrix, each row represents a bacterial isolate and each column represents a specific, variable genetic locus. The task then becomes a classic supervised machine learning problem: to build a model that can learn the complex, non-linear mapping from this high-dimensional genetic space to a specific phenotype.

A key feature of the SNP matrix is that the columns, representing genetic loci, are ordered according to their position on the reference chromosome. While this representation discards the non-variant nucleotides between SNPs, it preserves the crucial \textit{relative ordering and spatial context} of the variations themselves. This ordered series of SNPs can be treated as a new, compressed \textit{sequence of variations}. This perspective is biologically significant because the function of a gene is often determined not by a single point mutation, but by the collective effect of multiple SNPs acting in concert within a localized region\cite{Permana2023GraphSNP},~\cite{Shi2019DeepFeature}. Therefore, models that can process this data sequentially have the potential to uncover patterns that would be lost in a simple "bag-of-features" analysis.

Tree-based methods, such as Random Forest~\cite{Breiman2001RF} and XGBoost~\cite{Chen2016XGBoost}, serve as powerful baselines for this task. These models operate on the SNP matrix as a "bag of features," making decisions by recursively partitioning the data based on the most informative SNPs. Their strength lies in their ability to capture complex interactions between features, regardless of their position in the genome. However, their inherent disregard for the sequential order of SNPs means they may overlook crucial information encoded in the local context of mutations, such as multiple, functionally related SNPs occurring within the same gene.

To leverage this sequential information, models from deep learning offer a compelling alternative~\cite{Ji2021DNABERT},~\cite{Shi2019DeepFeature},~\cite{Ren2022},~\cite{siddiqui2025audiofuse}. A 1D Convolutional Neural Network (CNN) is particularly well-suited for this type of data~\cite{vandenBelt2024FASTNN},~\cite{Sharma2025Encoding}. By applying convolutional filters that slide across the ordered sequence of SNPs, a 1D CNN can learn to recognize short, recurring "motifs", local patterns of variation that are predictive of resistance. By stacking multiple convolutional and pooling layers, the model can learn a hierarchical representation of features, combining simple motifs into more complex patterns. This approach introduces a powerful \textit{inductive bias} that prioritizes local genomic context, a feature that is biologically relevant but absent in traditional "bag-of-features" models. The final component of our framework is a Multi-Layer Perceptron (MLP) head, which serves as the classifier on top of the features extracted by the CNN body, learning the final, non-linear decision boundary required to map the learned features to a resistance probability.

\section{Methodology}

\subsection{Dataset and Feature Engineering}
This study utilized the publicly available GieSSen dataset, a benchmark resource for AMR prediction originally curated by Ren et al. \cite{Ren2022}. The dataset consists of 809 \textit{E. coli} isolates, each with a corresponding pan-genomic SNP profile and experimentally determined antimicrobial susceptibility phenotypes.

The genomic data is provided as a SNP-matrix, constructed by aligning each isolate's genome to the \textit{E. coli} K-12 MG1555 reference strain and calling variants. This process results in a matrix where each row corresponds to a bacterial isolate and each column represents a specific SNP locus ordered by its chromosomal position. The nucleotide at each locus (A, C, G, T) and non-variant sites (N) were numerically encoded as integer tokens from 0 to 4. This encoding preserves the relative sequence of genetic variations, forming the basis for our sequence-aware models. The final matrix used in this study contained 809 samples and 60,936 SNP features.

Phenotypic labels for four antibiotics: Ciprofloxacin (CIP), Cefotaxime (CTX), Ceftazidime (CTZ), and Gentamicin (GEN), were provided as binary outcomes (0 for Susceptible, 1 for Resistant). Table~\ref{tab:dataset_summary} provides a detailed summary of the class distribution for each of the four prediction tasks. As shown, the proportion of resistant isolates ranges from a nearly balanced 45.2\% for Ciprofloxacin to a highly imbalanced 23.2\% for Gentamicin.

\begin{table*}[h!]
\centering
\caption{\textbf{Summary of the AMR Dataset Composition for Each Antibiotic}}
\label{tab:dataset_summary}
\begin{tabular}{lcccc}
\toprule
\textbf{Antibiotic} & \textbf{Total Samples} & \textbf{Susceptible (Class 0)} & \textbf{Resistant (Class 1)} & \textbf{\% Resistant} \\
\midrule
Ciprofloxacin (CIP) & 809 & 443 & 366 & 45.24\% \\
Cefotaxime (CTX) & 809 & 451 & 358 & 44.25\% \\
Ceftazidime (CTZ) & 809 & 533 & 276 & 34.12\% \\
Gentamicin (GEN) & 809 & 621 & 188 & 23.24\% \\
\bottomrule
\end{tabular}
\end{table*}

\subsection{Model Architectures}
We developed and compared three distinct modeling approaches: a deep 1D Convolutional Neural Network (CNN), a Random Forest baseline, and an XGBoost baseline. Finally, we created \textbf{AMR-EnsembleNet}, a soft voting ensemble that combines the outputs of our two best-performing models: the 1D CNN and XGBoost, one of which is sequence-aware and the other feature-based, making them complementary.

\subsubsection{1D Convolutional Neural Network (CNN)}
To effectively capture local sequence context from the SNP data, we designed a lightweight, deep 1D CNN. The architecture shown in Figure~\ref{fig:cnn_architecture} was extensively tuned for this specific task and is structured as follows:

\begin{figure*}[h!]
    \centering
    \includegraphics[width=0.95\textwidth]{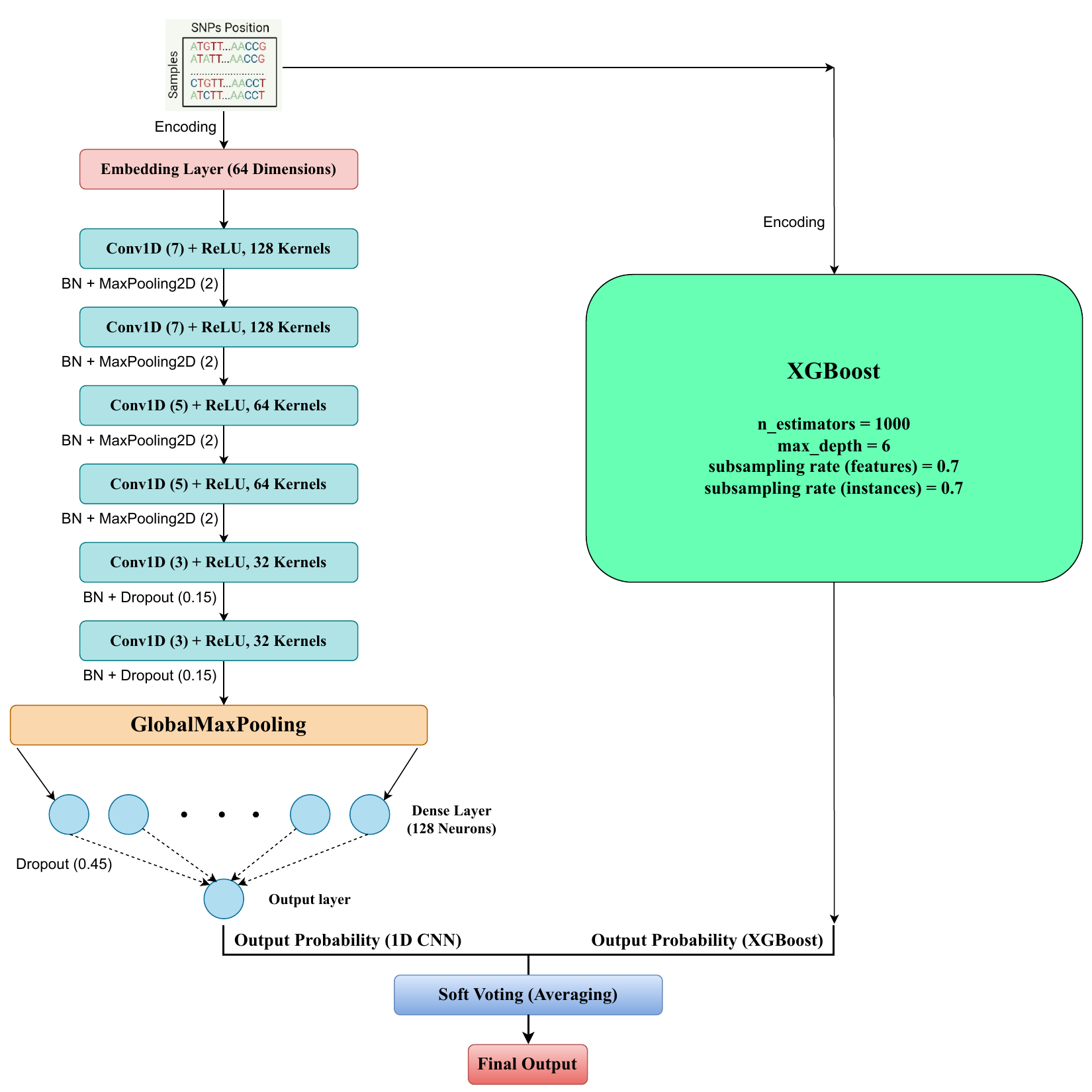}
    \caption{
        Architecture of \textbf{AMR-EnsembleNet}. In the 1D-CNN branch, an embedding layer processes the input SNP sequence, followed by six convolutional blocks that extract hierarchical features. A global max pooling layer and an MLP head produce the final resistance prediction. The predicted probabilities of this 1D-CNN branch are finally ensembled with XGBoost ones.
    }
    \label{fig:cnn_architecture}
\end{figure*}

\begin{enumerate}
    \item \textbf{Embedding Layer:} The input integer-token sequence is first passed through an embedding layer, which maps each discrete nucleotide token into a dense 64-dimensional vector representation. WE posit that embedding dimensions need not be too high as there are only 5 possible tokens.We posit that a high-dimensional embedding is unnecessary for this low-cardinality vocabulary. A moderately-sized dimension of 64 is sufficient for the model to learn meaningful relationships between the nucleotides (e.g., purines vs. pyrimidines) without introducing an excessive number of parameters that could lead to overfitting.
    
    \item \textbf{Hierarchical Convolutional Blocks:} The core of the model consists of six convolutional blocks designed to learn hierarchical features. Early blocks use larger filters (128) and kernel sizes (7x1) to detect broader sequence motifs, while deeper blocks use progressively smaller filters (64 and 32) and kernel sizes (5x1 and 3x1) to learn finer-grained patterns. Each block consists of a \texttt{Conv1D} layer, followed by \texttt{BatchNormalization} and a \texttt{ReLU} activation. \texttt{MaxPooling1D} layers with a pool size of 2 are strategically placed after the first four blocks to downsample the sequence and increase the receptive field of later layers. The kernel sizes for the convolutional layers were deliberately chosen to be relatively small (ranging from 3 to 7). While the input SNP sequence is very long, the fundamental unit of information is often the single nucleotide. We hypothesized that using large kernels could dilute the signal of a single, critical SNP by averaging it with its less informative neighbors. Instead, our deep, hierarchical architecture is designed to first identify short, significant SNP motifs with smaller kernels and then learn the more complex, longer-range patterns by combining these motifs in the deeper layers of the network.
    
    \item \textbf{Global MaxPooling:} After the final convolutional block, a \texttt{GlobalMaxPooling1D} layer was applied to aggregate features across the sequence dimension. We selected max pooling over average pooling based on empirical results and a key hypothesis about the nature of AMR. We posited that resistance phenotypes are often driven by sparse but high-impact genetic events, such as a single critical SNP or the presence of a specific resistance gene motif. \texttt{GlobalMaxPooling1D} is particularly effective in such scenarios, as it identifies the presence of the most salient feature detected anywhere in the genome, whereas average pooling would dilute this strong, localized signal with noise from non-informative regions.
    
    \item \textbf{MLP Classification Head:} This summary vector is passed to a Multi-Layer Perceptron (MLP) head for final classification. It consists of a dense hidden layer with 128 units and a \texttt{ReLU} activation, followed by a final output layer with a single neuron and a \texttt{sigmoid} activation to produce the probability of resistance. Dropout and L2 regularization were applied throughout the network to mitigate overfitting.
\end{enumerate}

\subsubsection{Baseline Models}
To contextualize the performance of our deep learning approach, we implemented two state-of-the-art tree-based models for tabular data:
\begin{itemize}
    \item \textbf{Random Forest:} A classifier consisting of 400 decision trees was trained. To handle the high dimensionality, the number of features considered at each split was limited to the square root of the total number of SNPs, and the maximum depth of each tree was capped at 15.
    
    \item \textbf{XGBoost:} An XGBoost classifier was implemented. The model was configured with a learning rate of 0.05, a maximum depth of 6, and subsampling of both instances (70\%) and features (70\%) per tree to ensure robustness.
\end{itemize}

\subsection{Ensemble Model}
To create a more robust predictor, we developed \textbf{AMR-EnsembleNet}, a soft voting ensemble that combines the outputs of our two best-performing models: the 1D CNN and XGBoost. The final prediction is calculated as the unweighted average of the probability scores generated by each of the two base models. This approach leverages the complementary strengths of the sequence-aware CNN and the feature-interaction-focused XGBoost model without requiring additional training of a meta-classifier. Figure~\ref{fig:cnn_architecture} portrays the overall architecture of \textbf{AMR-EnsembleNet}.

\subsection{Training and Evaluation Protocol}
For each antibiotic, the dataset was split into a training set (80\%) and a held-out test set (20\%) using stratified sampling to preserve the class distribution. For the deep learning and XGBoost models, the training set was further subdivided to create a validation set for initial hyperparameter tuning.

To address the significant class imbalance present in the CTX, CTZ, and GEN datasets, we employed a class weighting strategy during model training. The weights were calculated as inversely proportional to the roots of the class frequencies in the training data, thereby assigning a higher penalty to misclassifications of the minority (resistant) class.

Model performance was assessed on the held-out test set using a comprehensive suite of metrics: Accuracy, Matthews Correlation Coefficient (MCC), Cohen's Kappa, Macro F1-score, and the precision, recall, and F1-score for the resistant class. All reported metrics are formatted to four decimal places.

\section{Results and Discussion}
We evaluated our proposed 1D Convolutional Neural Network (CNN), the baseline Random Forest and XGBoost models, and our final 1D CNN-XGBoost ensemble on their ability to predict resistance to four antibiotics: Ciprofloxacin (CIP), Cefotaxime (CTX), Ceftazidime (CTZ), and Gentamicin (GEN). The performance of each model on the held-out test set is detailed below, focusing on Accuracy, Matthews Correlation Coefficient (MCC), and Precision, Recall, and F1-score for the resistant class, alongside the Macro F1-score for overall class balance.

\subsection{Performance on Ciprofloxacin (CIP) Resistance Prediction}
Ciprofloxacin represents a case with a relatively balanced class distribution and strong, well-characterized genetic determinants of resistance. As shown in Table~\ref{tab:results_cip}, all models achieved excellent predictive performance. The XGBoost model and our proposed ensemble delivered the highest accuracy (0.9630) and Matthews Correlation Coefficient (0.9253 and 0.9260, respectively). The strong performance across all models validates the underlying quality of the SNP data and confirms that for phenotypes with clear genetic signals, multiple advanced machine learning methods can achieve high accuracy, even though \textbf{AMR-EnsembleNet} performs marginally better than other models.

\begin{table*}[h!]
\centering
\caption{\textbf{Model Performance for Ciprofloxacin (CIP) Resistance}}
\label{tab:results_cip}
\begin{tabular}{lcccccc}
\toprule
\textbf{Model} & \textbf{Accuracy} & \textbf{F1 Score} & \textbf{Matthews} & \textbf{Precision} & \textbf{Recall} & \textbf{F1 Score (Macro)} \\
\midrule
Random Forest & 0.9568 & 0.9517 & 0.9127 & 0.9583 & 0.9452 & 0.9563 \\
XGBoost & \textbf{0.9630} & \textbf{0.9583} & 0.9253 & 0.9718 & 0.9452 & \textbf{0.9625} \\
1D CNN & 0.9568 & 0.9524 & 0.9129 & 0.9459 & \textbf{0.9589} & 0.9564 \\
\textbf{AMR-EnsembleNet} & \textbf{0.9630} & 0.9577 & \textbf{0.9260} & \textbf{0.9855} & 0.9315 & 0.9624 \\
\bottomrule
\end{tabular}
\end{table*}

\subsection{Performance on Moderately Imbalanced Datasets (CTX \& CTZ)}
The datasets for CTX and CTZ presented a greater challenge due to moderate class imbalance. For CTX (Table~\ref{tab:results_ctx}), \textbf{AMR-EnsembleNet} demonstrated clear performance improvements, emerging as the best model across all key metrics, including Accuracy (0.8086), F1 Score (0.8000), MCC (0.6241), and Macro F1-score (0.8083). Most notably, the ensemble achieved a resistant-class recall of 0.8611, a significant improvement over all other models, showcasing its enhanced ability to correctly identify true resistant isolates. This is a critical outcome for clinical applications.

\begin{table*}[h!]
\centering
\caption{\textbf{Model Performance for Cefotaxime (CTX) Resistance}}
\label{tab:results_ctx}
\begin{tabular}{lcccccc}
\toprule
\textbf{Model} & \textbf{Accuracy} & \textbf{F1 Score} & \textbf{Matthews} & \textbf{Precision} & \textbf{Recall} & \textbf{F1 Score (Macro)} \\
\midrule
Random Forest & 0.8025 & 0.7867 & 0.6050 & \textbf{0.7564} & 0.8194 & 0.8014 \\
XGBoost & 0.8025 & 0.7867 & 0.6050 & \textbf{0.7564} & 0.8194 & 0.8014 \\
1D CNN & 0.7840 & 0.7619 & 0.5647 & 0.7467 & 0.7778 & 0.7821 \\
\textbf{AMR-EnsembleNet} & \textbf{0.8086} & \textbf{0.8000} & \textbf{0.6241} & 0.7467 & \textbf{0.8611} & \textbf{0.8083} \\
\bottomrule
\end{tabular}
\end{table*}

The results for CTZ resistance (Table~\ref{tab:results_ctz}) highlight a more complex predictive challenge. Here, the Random Forest baseline achieved the highest scores across all metrics. This suggests that the genetic signals for CTZ resistance may be captured particularly well by the specific feature interaction and decision boundary characteristics of the Random Forest algorithm. While \textbf{AMR-EnsembleNet} did not surpass the baseline in this specific case, it's important to note its role in mitigating the weaknesses of its constituent models. The standalone 1D CNN, for example, had the lowest recall (0.6727), while the XGBoost had the lowest precision (0.6842). Our ensemble achieved a more balanced profile, improving upon the CNN's recall and the XGBoost model's precision by 3\~4\% each, resulting in the second-highest MCC (0.5715). This demonstrates the ensemble's value in providing a stable, well-balanced performance even when it is not the top performer. More importantly, even though we selected XGBoost as our primary feature-based baseline due to its consistency, these results underscore the value of model diversity in an ensemble. The distinct performance profiles suggest that the choice of constituent models is a critical factor for achieving good generalization.

\begin{table*}[h!]
\centering
\caption{\textbf{Model Performance for Ceftazidime (CTZ) Resistance}}
\label{tab:results_ctz}
\begin{tabular}{lcccccc}
\toprule
\textbf{Model} & \textbf{Accuracy} & \textbf{F1 Score} & \textbf{Matthews} & \textbf{Precision} & \textbf{Recall} & \textbf{F1 Score (Macro)} \\
\midrule
Random Forest & \textbf{0.8210} & \textbf{0.7290} & \textbf{0.5960} & \textbf{0.7500} & \textbf{0.7901} & \textbf{0.7977} \\
XGBoost & 0.7901 & 0.6964 & 0.5363 & 0.6842 & 0.7091 & 0.7680 \\
1D CNN & 0.8086 & 0.7048 & 0.5651 & 0.7400 & 0.6727 & 0.7816 \\
\textbf{AMR-EnsembleNet} & 0.8086 & 0.7156 & 0.5715 & 0.7222 & 0.7091 & 0.7857 \\
\bottomrule
\end{tabular}
\end{table*}

\subsection{Performance on the Highly Imbalanced Gentamicin (GEN) Dataset}
The Gentamicin dataset served as the most stringent test of model robustness due to the severe class imbalance. XGBoost's high accuracy is misleading here, as is evident from the lower F1 Score, which is crucial for an imbalanced dataset. The results (Table~\ref{tab:results_gen}) revealed the unique strength of the sequence-aware 1D CNN. It achieved a recall of 0.7105 for the resistant class, dramatically outperforming both Random Forest (0.5263) and XGBoost (0.5789). However, this high recall came at the cost of the lowest precision (0.4576). The XGBoost model, conversely, achieved the highest precision (0.4889) but with a much lower recall.

This is where the power of the ensemble becomes most evident. Our \textbf{AMR-EnsembleNet} successfully capitalized on the strengths of both models, achieving a recall (0.6842) nearly as high as the CNN's, while simultaneously improving upon its precision. This balancing act resulted in the ensemble achieving the highest performance on the metrics that evaluate the complete classification picture: the Matthews Correlation Coefficient (0.4030) and the Macro F1-score (0.6908). It effectively synthesized the CNN's ability to find rare resistant cases with the XGBoost's more conservative decision-making, leading to the most balanced and reliable classifier for this difficult task.

\begin{table*}[h!]
\centering
\caption{\textbf{Model Performance for Gentamicin (GEN) Resistance}}
\label{tab:results_gen}
\begin{tabular}{lcccccc}
\toprule
\textbf{Model} & \textbf{Accuracy} & \textbf{F1 Score} & \textbf{Matthews} & \textbf{Precision} & \textbf{Recall} & \textbf{F1 Score (Macro)} \\
\midrule
Random Forest & 0.7469 & 0.4938 & 0.3271 & 0.4651 & 0.5263 & 0.6626 \\
XGBoost & \textbf{0.7593} & 0.5301 & 0.3722 & \textbf{0.4889} & 0.5789 & 0.6841 \\
1D CNN & 0.7346 & 0.5567 & 0.3984 & 0.4576 & \textbf{0.7105} & 0.6836 \\
\textbf{AMR-EnsembleNet} & 0.7469 & \textbf{0.5591} & \textbf{0.4030} & 0.4727 & 0.6842 & \textbf{0.6908} \\
\bottomrule
\end{tabular}
\end{table*}

\subsection{Synergistic Advantage of the Ensemble Framework}
A synthesis of the results across all four antibiotics reveals a consistent and powerful narrative. No single model was universally superior; instead, the 1D CNN and XGBoost demonstrated complementary strengths rooted in their distinct architectural biases. The XGBoost model, treating SNPs as an unordered set of features, excelled when the predictive signal was strong (CIP). In contrast, the 1D CNN, processing SNPs as an ordered sequence, proved uniquely adept at identifying true resistant isolates (high recall) when the signal was sparse and the data highly imbalanced (GEN).

Our \textbf{AMR-EnsembleNet} consistently provided the most robust generalization across these diverse challenges. For CTX and GEN, it achieved the best overall performance by creating a classifier superior to its individual components. For CIP, it matched the top performance. And for CTZ, where a different baseline performed best, the ensemble acted as a stabilizer, mitigating the individual weaknesses of the CNN and XGBoost to produce a strong, balanced result. It leverages the high recall of the CNN in data-scarce situations and the high precision of XGBoost in high-signal situations. By fusing a model that understands \textit{what} SNPs are present with a model that understands \textit{where} they are located, the ensemble framework consistently delivers a reliable, high-performing, and well-balanced solution, making it an ideal strategy for the complex and varied nature of genomic AMR prediction.

\section{Model Interpretability and Biological Validation}

A key objective of this study was not only to predict AMR but also to understand the genetic basis for those predictions. To achieve this, we employed SHAP (SHapley Additive exPlanations)~\cite{Lundberg2017SHAP} to interpret our best-performing XGBoost models for each of the four antibiotics. SHAP analysis assigns an importance value to each SNP for every individual prediction, allowing us to identify the specific genetic loci that are the most influential drivers of the model's decisions.

% --- Figure for 2x2 SHAP Summary Plots ---
\begin{figure*}[h!]
    \centering

    % --- Top Row ---
    \begin{minipage}{0.48\textwidth}
        \centering
        \includegraphics[width=\linewidth]{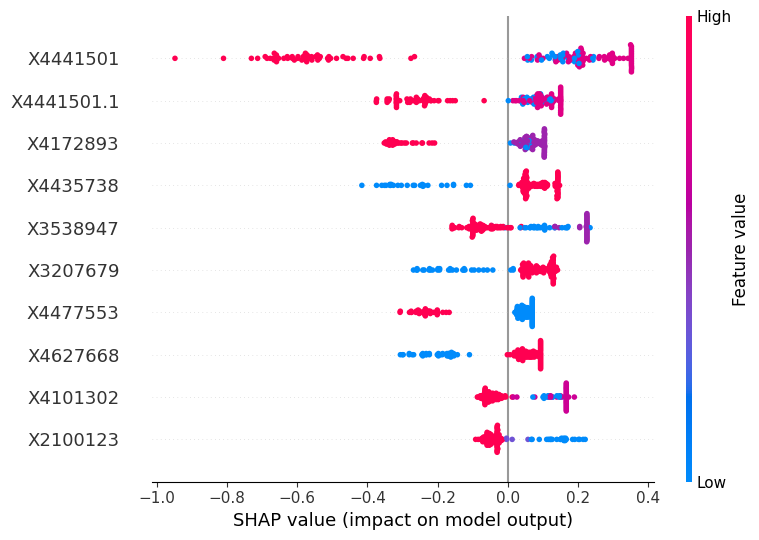}
        \label{fig:shap_cip}
        \centerline{(A) Ciprofloxacin (CIP)}
    \end{minipage}\hfill
    \begin{minipage}{0.48\textwidth}
        \centering
        \includegraphics[width=\linewidth]{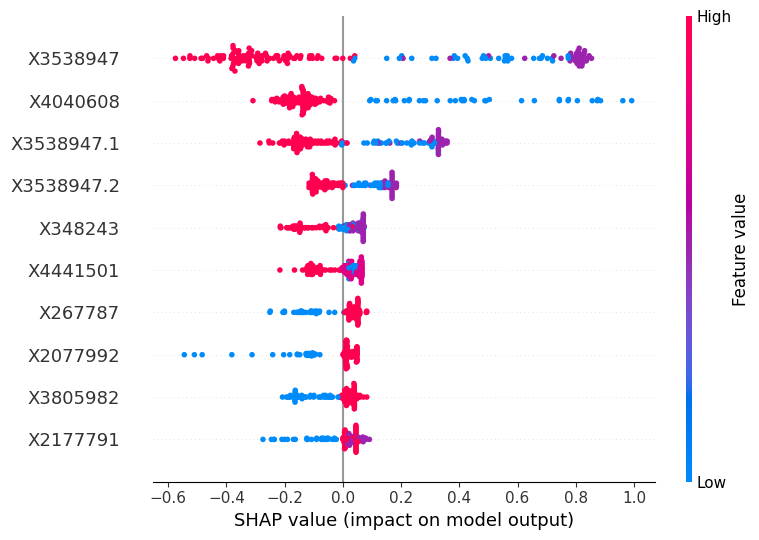}
        \label{fig:shap_ctx}
        \centerline{(B) Cefotaxime (CTX)}
    \end{minipage}

    \vspace{0.5cm} 

    % --- Bottom Row ---
    \begin{minipage}{0.48\textwidth}
        \centering
        \includegraphics[width=\linewidth]{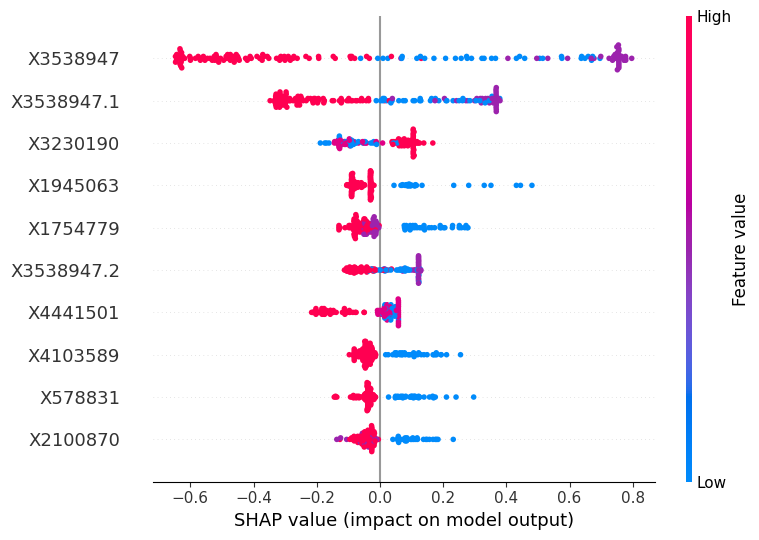}
        \label{fig:shap_ctz}
        \centerline{(C) Ceftazidime (CTZ)}
    \end{minipage}\hfill
    \begin{minipage}{0.48\textwidth}
        \centering
        \includegraphics[width=\linewidth]{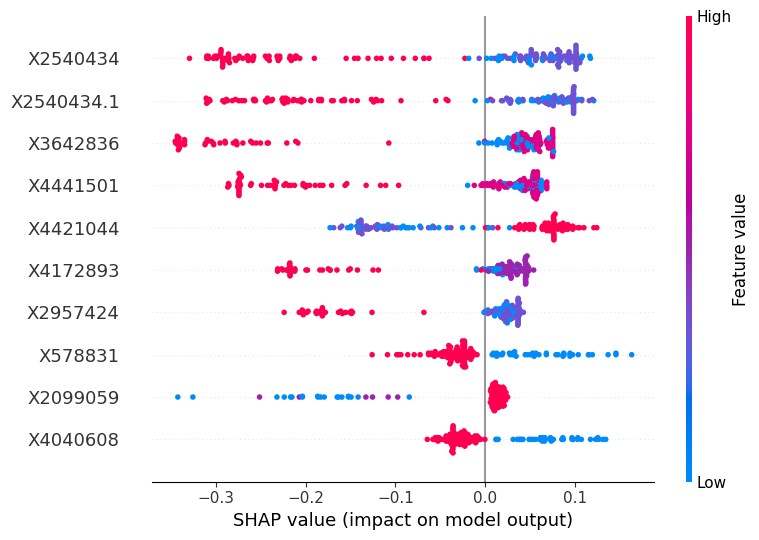}
        \label{fig:shap_gen}
        \centerline{(D) Gentamicin (GEN)}
    \end{minipage}

    \caption{SHAP summary plots showing the top-ranked impactful SNP features for each of the four antibiotics. Each plot displays features ranked by their mean absolute SHAP value. For each feature, a single dot represents a sample from the test set. The horizontal position of the dot shows its SHAP value (impact on the model output), with positive values pushing the prediction towards "Resistant". The color indicates the original feature value (red=high, blue=low).}
    \label{fig:shap_summary_grid}
\end{figure*}

\subsection{Identifying Key Predictive SNPs with SHAP}
For each antibiotic, we generated a SHAP summary plot (Figure~\ref{fig:shap_summary_grid}) to visualize the top 10 most impactful SNP features. These plots reveal the direction and magnitude of each SNP's effect on the resistance prediction. For example, in the prediction of Ciprofloxacin (CIP) resistance, the SNP at chromosomal position 4,435,738 (X4435738) was identified as one of the most influential feature. The plot clearly shows that a high feature value (corresponding to a `T` or `C` nucleotide) at this locus pushes the model's output towards predicting resistance (positive SHAP value), while a low feature value (`A` or `N`) pushes the prediction towards susceptibility (negative SHAP value).

This pattern of a few high-impact SNPs, complemented by a larger set of features with smaller, consistent effects, was observed across all antibiotics. The SHAP plots clearly demonstrate the polygenic nature of the model's predictions. Resistance is not determined by a single feature, but by the cumulative and sometimes opposing impacts of multiple SNPs. For instance, in the Gentamicin (GEN) model, SNPs at positions 2,099,059 and 4,421,044 both contribute positively to resistance predictions when their feature values are high, showcasing a multi-locus genetic architecture learned by the model.

\subsection{Biological Validation of Top-Ranked SNPs}
To validate the biological relevance of our model's findings, we mapped the chromosomal positions of the top-ranked SNPs to the \textit{E. coli} K-12 MG1655 reference genome~\cite{Blattner1997},~\cite{NCBI_GGD}. This analysis confirmed that our model independently rediscovered several of the most critical and well-documented genes associated with AMR in \textit{E. coli}.

\begin{table*}[h!]
\centering
\caption{\textbf{Biological Validation of Top SNPs Identified by SHAP Analysis}}
\label{tab:shap_validation}
\begin{tabularx}{\textwidth}{l c c >{\RaggedRight}X}
\toprule
\textbf{Antibiotic} & \textbf{SNP Position} & \textbf{\begin{tabular}[c]{@{}c@{}}Associated\\ Gene\end{tabular}} & \textbf{Known Function / Relevance to AMR} \\
\midrule
Ciprofloxacin & X4441501 & \textit{gyrB} & DNA gyrase subunit B; a known secondary target for quinolone resistance~\cite{Hooper2001}. \\
Ciprofloxacin & X4172893 & \textit{marR} & Transcriptional repressor of the AcrAB-TolC multi-drug efflux pump~\cite{Redgrave2014}. \\
\midrule
Gentamicin & X2540434 & \textit{rpsL} & Ribosomal protein S12; a canonical target for aminoglycoside resistance~\cite{Jana2006}. \\
Gentamicin & X4421044 & \textit{parC} & Topoisomerase IV subunit A; its mutation may indicate co-resistance or affect cell fitness~\cite{Redgrave2014}. \\
\midrule
Cefotaxime & X3538947 & \textit{fusA} & Elongation factor G; its high rank for a beta-lactam suggests a potential pleiotropic effect~\cite{Jana2006}. \\ 
Cefotaxime & X4040608 & \textit{ampC} & Beta-lactamase; mutations in its promoter region can increase expression and cause resistance~\cite{Jacoby2009}. \\ 
\midrule
Ceftazidime & X3538947 & \textit{fusA} & Elongation factor G; its repeated high importance suggests a role in general antibiotic stress response~\cite{Jana2006}. \\ 
Ceftazidime & X578831 & \textit{acrD} & A component of the AcrAD-TolC multi-drug resistance efflux pump~\cite{Poole2004}. \\ 
Ceftazidime & X3230190 & \textit{ftsI} (PBP3) & Penicillin-binding protein 3; a primary target for beta-lactam antibiotics~\cite{Sauvage2008}. \\ 
\bottomrule
\end{tabularx}
\end{table*}

A selection of these key findings is presented in Table~\ref{tab:shap_validation}. For Gentamicin, an aminoglycoside antibiotic, the model's most influential feature was the SNP at position 2,540,434 (X2540434). This locus is located directly within the \textbf{\textit{rpsL}} gene, which encodes the ribosomal protein S12. Mutations in this gene, particularly at codon 43 (e.g., K43N), are the canonical and most famous mechanism of high-level resistance to streptomycin, another aminoglycoside antibiotic. Another top predictive SNP, X4421044, fall directly within the Quinolone Resistance-Determining Regions (QRDR) of the \textbf{\textit{parC}} gene~\cite{Eaves2004QRDR}. This gene encodes the primary targets of fluoroquinolone antibiotics, and mutations in these regions are the canonical mechanism of high-level resistance. Similarly, for Ciprofloxacin, the top predictive SNP was found at position 4,441,501 (X4441501), which maps to the \textbf{\textit{gyrB}} gene. This gene encodes a subunit of DNA gyrase, a primary target of fluoroquinolone antibiotics. While mutations in its partner gene, \textit{gyrA}, are more common, variations in \textit{gyrB} are a known and significant contributor to resistance.

Intriguingly, for both Cefotaxime (CTX) and Ceftazidime (CTZ), the model most heavily weighted the SNP at position 3,538,947 (X3538947), located in the \textbf{\textit{fusA}} gene. While \textit{fusA} is typically associated with aminoglycoside resistance~\cite{Holley2022fusA}, its consistent high ranking for these beta-lactam antibiotics suggests our model may be capturing complex, pleiotropic effects or signals of co-occurring resistance.

The ability of our model to identify these and other relevant loci (Table~\ref{tab:shap_validation}) without any prior biological programming serves as a powerful validation of our approach. It demonstrates that the framework is not merely fitting statistical noise, but is successfully learning the true underlying genetic architecture of the AMR phenotype. This establishes our method as a robust tool for both high-accuracy prediction and data-driven biological discovery.

\section{Conclusion}

In this study, we addressed the urgent challenge of antimicrobial resistance prediction by moving beyond the limitations of existing computational methods. We demonstrated that for genomic data, there is a powerful middle ground between order-agnostic machine learning models and data-intensive Transformers. Our work shows that a lightweight 1D Convolutional Neural Network can successfully learn the dynamics of the genome, extracting predictive patterns from the sequential context of SNPs. This architectural choice proved its merit on a highly imbalanced dataset, where the CNN's ability to recognize local motifs allowed it to identify resistant isolates (recall) far more effectively than traditional baselines.

The core contribution of our work is the demonstration of AMR-EnsembleNet, a synergistic ensemble framework. By fusing the sequence-aware 1D CNN with a feature-based XGBoost model, we created a classifier that is consistently more robust and balanced than either of its components alone. This ensemble excels by leveraging the complementary strengths of its constituents: the CNN's sensitivity to sparse, local signals and the XGBoost model's power in capturing complex, non-local feature interactions. Through model interpretation, we confirmed that this high performance is not a statistical artifact but is driven by the identification of SNPs in biologically crucial, known AMR-associated genes.

While this framework provides a robust and computationally feasible solution for AMR prediction, it also opens several exciting avenues for future research. A natural next step is to adapt the model for regression to predict quantitative Minimum Inhibitory Concentration (MIC) values, offering a more clinically nuanced output. Other data modalities can also be integrated, such as gene expression data (transcriptomics). Also, applying more advanced explainability methods to the CNN itself could move the framework beyond validating known biology and toward becoming an effective tool for discovering novel genetic markers of resistance.

Ultimately, our ensemble approach offers a powerful and practical tool for genomic surveillance, representing a significant step towards the future of rapid, data-driven diagnostics in the fight against antimicrobial resistance.

\section{Code and Data Availability}
All Python scripts, Jupyter notebooks, and configuration files used to perform the data preprocessing, model training, evaluation, and ensembling described in this paper are publicly available on GitHub at: \url{https://github.com/Saiful185/AMR-EnsembleNet}.

The dataset used in this study is a publicly available benchmark resource. The pan-genomic SNP-matrix and the corresponding AMR phenotype data for the 809 \textit{E. coli} isolates were obtained from the ML-iAMR GitHub repository, originally curated from the GieSSen dataset \cite{Ren2022}.

\begin{acks}
The authors wish to express sincere gratitude to the organizers and mentors of the CompBioAsia 2025 workshop, where this project was initiated. Special thanks are extended to Dr. Travis Wheeler and Jeremiah Gaiser of the University of Arizona for their invaluable guidance, insightful discussions, and technical support throughout the early stages of this research. The authors are also deeply grateful to Elizabeth Nemeti of Emory University for her critical feedback, which significantly shaped the direction of this work.
\end{acks}

\bibliographystyle{ACM-Reference-Format}
\bibliography{base}


\end{document}